\documentclass[final]{cvpr}

\usepackage{times}
\usepackage{epsfig}
\usepackage{graphicx}
\usepackage{amsmath}
\usepackage{amssymb}
\usepackage{physics}
\usepackage{enumitem}
\usepackage{bm}
\usepackage{microtype}

\usepackage[pagebackref,breaklinks,colorlinks]{hyperref}

\usepackage{bbm}
\usepackage{caption}

\title{SketchEdit: Image Editing with Partial Sketches}

\def\cvprPaperID{847} % *** Enter the CVPR Paper ID here
\def\confName{CVPR}
\def\confYear{2022}
\begin{document}

%%%%%%%%% TITLE
\title{SketchEdit: Mask-Free Local Image Manipulation with Partial Sketches}

\author{Yu Zeng\\
Johns Hopkins University\\
{\tt\small yzeng22@jhu.edu}
\and
Zhe Lin\\
Adobe Research\\
{\tt\small zlin@adobe.com}
\and
Vishal M. Patel\\
Johns Hopkins University\\
{\tt\small vpatel36@jhu.edu}
}

%%%%%%%%% TITLE
% \maketitle

\twocolumn[{%
\renewcommand\twocolumn[1][]{#1}%
\maketitle
\begin{center}
    \centering
    \includegraphics[width=\textwidth]{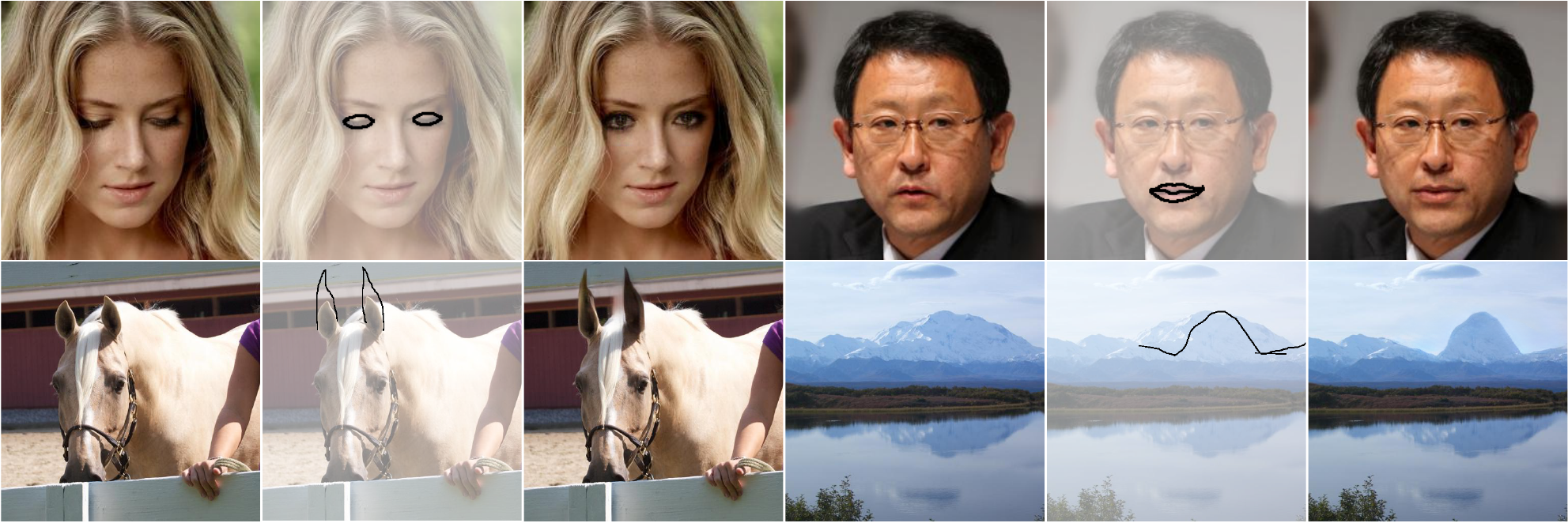}\\
    \scriptsize{\hfill {Images} \hfill\hfill  {Sketches} \hfill\hfill  {Results} \hfill\hfill {Images} \hfill\hfill {Sketches} \hfill\hfill {Results} \hfill}
    \vspace{0.30cm}
    \captionof{figure}{Our system allows users to interactively manipulate an image by sketching desired contours directly on top of the image. %\zl{column text not centered..., top-right image can be replaced with more dramatic one...bottom-right bird example is not very good, can we change one?}
    }
    \vspace{-0.3em}
\end{center}%
}]

\makeatother

%%%%%%%%% ABSTRACT

\begin{abstract}
Sketch-based image manipulation is an interactive image editing task to modify an image based on input sketches from users. Existing methods typically formulate this task as a conditional inpainting problem, which requires users to draw an extra mask indicating the region to modify in addition to sketches. 
The masked regions are regarded as holes and filled by an inpainting model conditioned on the sketch. 
With this formulation, paired training data can be easily obtained by randomly creating masks and extracting edges or contours . 
% Then the problem can be solved by training an inpainting model to predict the original image given the incomplete partial image, mask, and sketch as input. 
Although this setup simplifies data preparation and model design, it complicates user interaction and discards useful information in masked regions. 
To this end, we investigate a new paradigm of sketch-based image manipulation: mask-free local image manipulation, which  only requires sketch inputs from users and utilizes the entire original image. 
% To this end, we propose a new framework for sketch-based image manipulation that only requires sketch inputs from users and utilizes the entire original image. 
Given an image and sketch, our model automatically predicts the target modification region and encodes it into a structure agnostic style vector. 
A generator then synthesizes the new image content based on the style vector and sketch. 
The manipulated image is finally produced by blending the generator output into the modification region of the original image. 
Our model can be trained in a self-supervised fashion by learning the reconstruction of an image region from the style vector and sketch. 
The proposed method offers simpler and more intuitive user workflows for sketch-based image manipulation and provides better results than previous approaches. More results, code and interactive demo will be available at \url{https://zengxianyu.github.io/sketchedit}. 
%The proposed framework offers simpler and more intuitive user workflow, and achieves better results than previous approaches. The code and interactive demo can be found in the supplementary material. 
%, as supported by the experimental results.
\end{abstract}

%%%%%%%%% BODY TEXT
\section{Introduction}
Recently there have been increasing efforts and demand for building interactive photo editing tools on devices with touch interfaces.   
%tools among Internet and social media users to edit photos with an easy and intuitive interface. 
Being expressive and easily editable, sketching is one of the most straightforward ways that people illustrate their creative ideas and interact with the apps~\cite{chen2018sketchygan,wang2021sketch,ghosh2019interactive,liu2019sketchgan}. Sketch-based image editing is an emerging research topic where the goal is to build models which can manipulate holistic or local structures of an image according to the user-drawn sketches. It has made a significant progress in the last few years with advancements in deep learning and generative models, \eg, GANs~\cite{goodfellow2014generative}. 
%\eg, Generative Adversarial Networks (GANs)~\cite{goodfellow2014generative}. 

\begin{figure}[t]
\begin{center}
% \fbox{\rule{0pt}{1.5in} \rule{\linewidth}{0pt}}
\includegraphics[width=\linewidth]{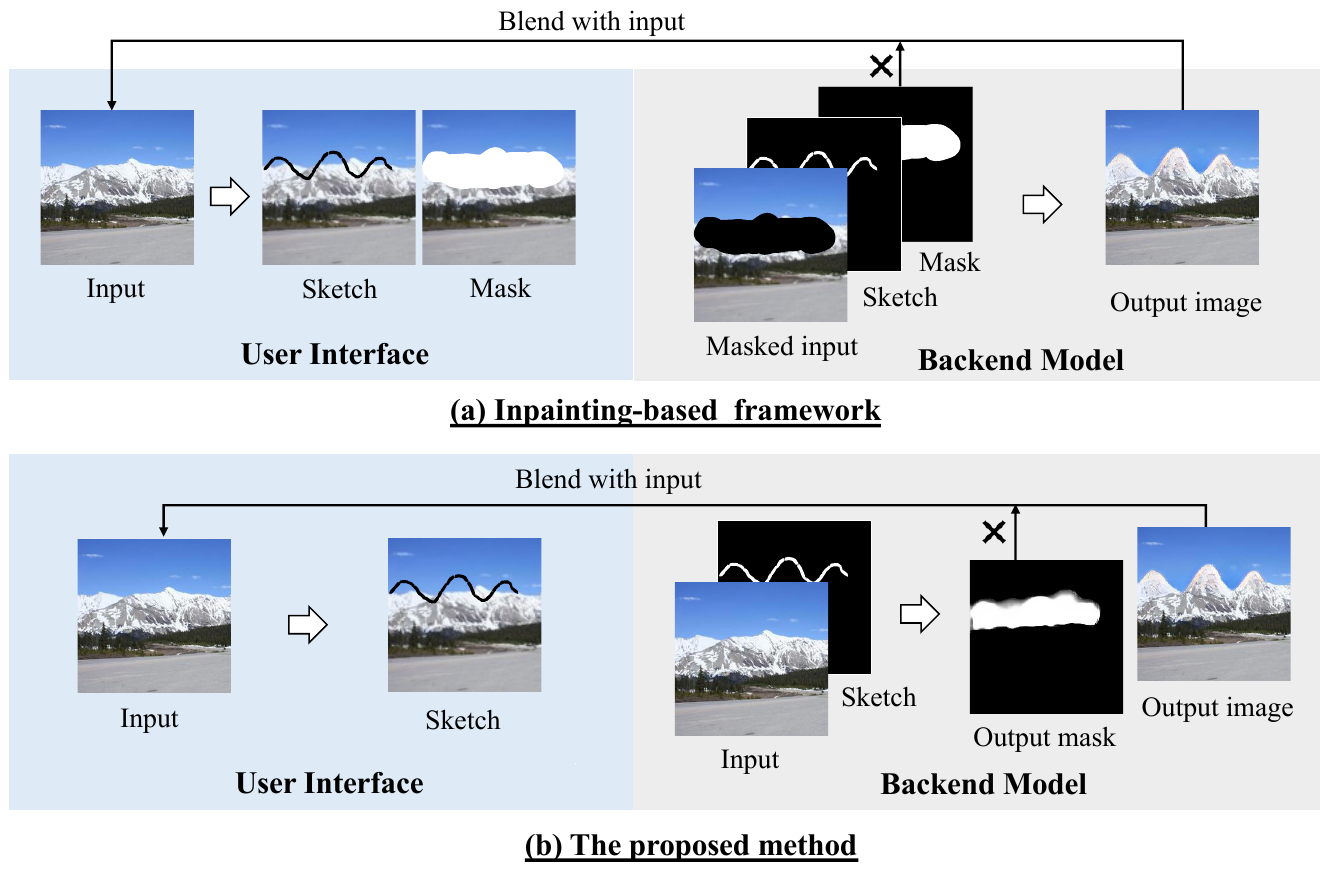}
\end{center}
% \vspace{-10pt}
\caption{Illustration of the previous inpainting-based image manipulation systems and the proposed one with only sketch input. 
%\zl{better to move figures to the top.}
}
\label{fig_userinter}
\end{figure}

There are two challenges in sketch-based image manipulation: (1) The input sketch roughly indicates where to modify, but the precise modification region is unknown, and (2) it is difficult to collect large-scale image pair data (\ie images before and after manipulation) for training. 
Previous approaches avoid these obstacles by converting sketch-based image manipulation into a conditional inpainting problem. 
% [[In addition to the sketch, users are required to draw an extra mask to indicate the modification regions, which are regarded as holes to be filled by an inpainting model conditioned on the sketch. ]]
Users are required to draw an extra mask to indicate the modification regions in addition to the sketch. The masked regions are regarded as holes and filled by an inpainting model conditioned on the sketch, as illustrated in Fig.~\ref{fig_userinter}~(a). 
Under this inpainting framework, paired training data can be easily obtained by randomly generating masks as in general image inpainting~\cite{lahiri2020prior,suin2021distillation,zhou2021transfill,yi2020contextual,nazeri2019edgeconnect,iizuka2017globally,liu2018image,xiong2019foreground,ren2019structureflow,liao2021image,xiao2019cisi,yu2020region,yang2020learning,yang2017high,wangimage,pathak2016context,song2018contextual,ren2019structureflow} and extracting edges/contours in the masked regions as surrogate sketches using edge detection algorithms~\cite{canny1986computational,he2019bi,xie2015holistically,deng2018learning,deng2020deep,su2021pixel,xu2018learning,liu2021pd,liu2020rethinking,liao2020guidance,wang2020vcnet}. Then the problem can be solved by training an inpainting model to predict the original image given the masked image, mask, and sketch as input. 
%For training, a random region in an image is set to missing, ~\ie erase all pixels, and the learning objective of is to fill the missing region based on the structure information from the sketch and style information from surrounding areas. The sketch data used in training can obtained automatically from contour detection algorithms. In inference, the users draw the sketch and an extra mask of the precise modification regions, which are set to missing to be inpainted by the model. 
%In training, they add deterioration to images by masking at random locations, and train a conditional inpainting model to reconstruct the original image given an incomplete image and a sketch of the missing parts. 
%When using an inpainting-based model for image editing, the users are asked to draw a sketch and an extra mask indicating the area to modify. Then the masked region is set as missing and filled by the inpainting model. 
This clever formulation simplifies model design and training setup, and has been explored extensively in the recent literature~\cite{liu2021deflocnet,yang2020deep,jo2019sc,portenier2018faceshop,yu2019free}.  

However, the inpainting framework is sub-optimal for sketch-based image manipulation.  First, although it eases model design and training, it leaves extra work to users and results in a complicated user  interface. After users sketch in an image region, they have to draw a mask again around the same location. This procedure is tedious and redundant. 
An untrained user may not be able to draw a good mask that exactly covers the desired modification area, making the model difficult to produce reasonable results. 
Second, with this framework, masked regions in an input image has to be removed to match the training condition of the inpainting model. 
However, since the original content in a modification region is usually highly correlated with the desired result, ignoring this information will degrade the quality of the manipulated image and lead to unwanted changes in the modification region. 

To this end, we investigate a new paradigm of sketch-based image manipulation, which requires only sketch inputs from users while leveraging the entire original image. 
Our system provides a more straightforward and user friendly interface for image manipulation: to solely sketch directly on top of the original image, as illustrated in Fig.~\ref{fig_userinter}~(b). Moreover, as the system takes the entire image as input, information in the modification region can be reserved, resulting in more consistent appearance (\eg color, texture) with the original content in the modification region. 
%This framework provides a more straightforward and robust way of image manipulation, meanwhile allows to preserve more information from input images.   

For local editing, it is desired to preserve most of the original image content and only modify the relevant image region surrounding the sketch input. Therefore, we first predict the modification region with a mask estimator and then synthesize new content inside with a generator. 
%we decompose our framework into two components: mask prediction and generator. 
% we formulate two sub-tasks and solve them with two components: 
% to predict the precise modification region with a mask estimator based on joint inference of an input image and sketch, and to synthesize new content in the modification region with a generator. 
The manipulated image is produced by blending the generator output into the original image using the predicted mask. 
To encourage the structure of the synthesized content to follow the sketch while only keeping the style of the original content, we encode the modification region of an input image into a structure agnostic style vector with a style encoder. The system can be trained in a self-supervised fashion by learning to reconstruct the target modification region based on the style vectors and sketches.

We evaluate our method on multiple datasets, including CelebAHQ, Places2, and newly constructed datasets SketchFace and SketchImg containing sketches and masks drawn by users. Extensive experiments demonstrate that the proposed method outperforms the state-of-the-art approaches. To show the advantage of our new framework in terms of user interaction, we include an interactive demo in the supplementary material. Our contributions are as follow,
\begin{itemize}[noitemsep]
\item Investigation of a new paradigm of sketch-based image manipulation: mask-free local image manipulation that requires only partial sketch inputs from users. 
\item The first system for mask-free sketch-based local image manipulation, including the network  architecture, data acquisition, and training strategy. 
\item Extensive experiments are conducted on multiple datasets to demonstrate the superiority of our approach over the related methods. 
\end{itemize}
% \begin{itemize}[noitemsep]
% \item The first system for mask-free local image manipulation that requires only partial sketch inputs from users. 
% \item A complete framework in the new paradigm, including the network architecture, data acquisition, and training strategy. 
% \item Extensive experiments are conducted to demonstrate the superiority of our approach over the related methods. 
% \end{itemize}

\section{Related Work}
\noindent \textbf{Sketch-guided image completion. } Previous research on sketch-based image manipulation is studied mainly under a conditional image inpainting framework. 
% Usually sketch-guided inpainting model learns to fill the missing region in an incomplete input image given the sketch in the missing region. 
% Enormous training data can be obtained by randomly remove a square or random region from images and the sketch maps can by generated by edge detection models. 
% When used for image manipulation, the incomplete input image can be constructed by asking the users to draw an extra mask indicating the area to edit. Then this area is taken as missing and to be filled by the inpainting model guided by sketch.
Yu~\etal~\cite{yu2019free} propose DeepFill-v2 which can perform sketch-guided image inpainting of general images as well as face images. The gated convolution layer is introduced to select useful features from the incomplete input dynamically. Portenier~\etal~\cite{portenier2018faceshop} propose FaceShop for sketch-based face images manipulation through conditional image completion. FaceShop allows users to modify the local shape and color in a face image by drawing the mask, sketch, and a few color strokes. Jo and Park~\cite{jo2019sc} explore face manipulation with sketch and color strokes through image completion. They use free-form masks and style loss to train the image completion model to handle irregular and possibly large missing areas. Yang~\etal~\cite{yang2020deep} focus on the adaptation of the face manipulation model trained on sketches generated by edge detection to the human-drawn sketches. A sketch refinement strategy is proposed, which first dilates then refines a user-drawn sketch to close to an edge detection result. Our work is inspired by these methods but focuses on the manipulation of complete images. We do not require the users to draw an extra mask to construct an incomplete image when editing a photo. Instead, the region to edit is automatically discovered and modified based on where and what they draw. 

\noindent \textbf{Image-to-image translation. } Image translation aims to learn the mapping between different image domains, \eg to generate images from semantic label maps, edges, \etc. Isola~\etal~\cite{isola2017image} propose an image-to-image translation framework, called Pix2Pix, to generate images from label maps or edge maps. Zhu~\etal~\cite{zhu2017unpaired} propose CycleGAN, which allows training an image translation model on unpaired data with a cycle consistency constraint. Park~\etal~\cite{park2019semantic} propose spatially-adaptive normalization for image generation conditioned on semantic layout, which modulates the activations using input semantic layouts to propagate semantic information throughout the network better.  Zhang~\etal~\cite{zhang2020cross} propose a framework for exemplar-based image translation with cross-domain correspondence by jointly learning cross-domain correspondence and image translation. When dealing with the edge input, image translation methods usually require a complete edge map and generate a brand new image. In contrast, our method focuses on local image manipulation with partial sketches as input. 

\noindent \textbf{Contour-based image manipulation} is another line of research where an image is edited in the contour domain. Approaches in this category typically encode an image into an edge-based representation and decode after users modify the edges. Elder~\etal~\cite{elder1998image} encode images with the brightness on edges and perform image editing operations (crop, paste, delete) in the contour domain. Dekel~\etal\cite{dekel2018sparse} propose to encode images with learned features at contour points to perform complex changes in the image domain by simple edits of contours. Vinker~\etal~\cite{vinker2020deep} explore deep image manipulation with a single training sample by extensive augmentation of the training sample through thin-plate splines (TPS) warping. A conditional generator trained on the augmented samples can map the edited edges and segmentation to the modified image. These methods require the contour representation to be complete from which the image can be faithfully reconstructed and assume that the users have basic photo editing skills to perform certain editing operations on a pre-computed edge map. In comparison, our method allows a broader range of users to manipulate an image more easily,~\ie by directly sketching on top of the input image. 

\section{Proposed Method}
We focus on local image manipulation based on free-form sketches. 
Given an input image $x$ and partial sketch $c$ indicating the target contour, our model modifies a local region of $x$ to match the sketch, resulting in a manipulated image $y$. 
Due to the lack of data for supervised training, we propose a self-supervised learning approach to jointly learn mask estimation and masked region reconstruction. 

\subsection{Model}
\label{sec:model}
\begin{figure*}[t]
\begin{center}
% \fbox{\rule{0pt}{1.5in} \rule{\linewidth}{0pt}}
\includegraphics[width=.9\linewidth]{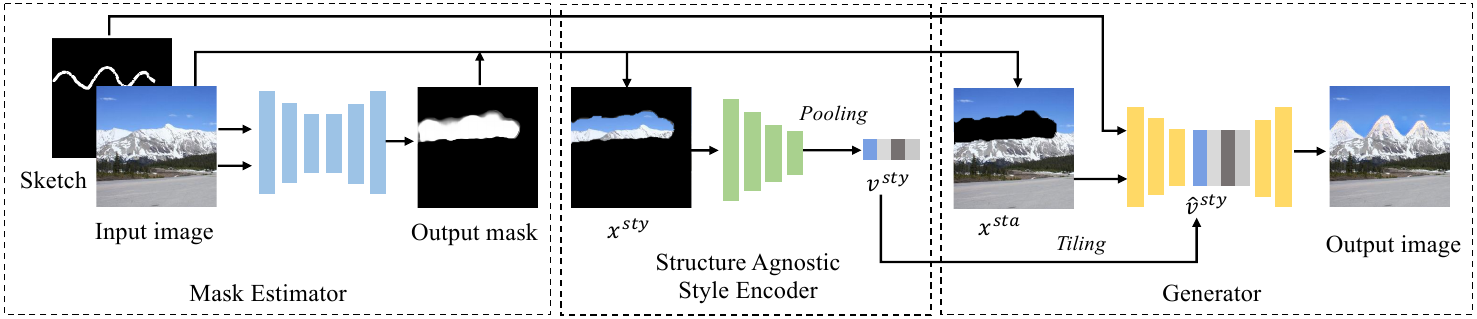}
\end{center}
% \vspace{-10pt}
\caption{Model structure of the proposed method. Given an input image $x$ and sketch $c$, the mask estimator predicts a mask $m$ indicating the region to edit. The style partial image $x^{sty}$ and the static partial image $x^{sta}$ can be created by multiplying $x$ with $m$ and $1-m$ element-wise, respectively. Then the generator synthesizes new content in the modification region based on the static image $x^{sta}$, sketch $c$, and a structure agnostic style feature extracted from $x^{sty}$ by the style encoder. 
%\zl{ZL: I feel it could be better by drawing it horizontally. Also, it is still better to have a bit more info on the caption.}
}
\label{fig_model}
\end{figure*}

Our model consists of three components: mask estimator, structure agnostic style encoder, and generator. 
Fig.~\ref{fig_model} shows the overall structure of our model. 
%The generator and mask estimator that synthesizes new content and produces a mask to blend the synthesized content into the original image.  
Given an input image $x$ and sketch $c$, the generator synthesizes new content based on the image, sketch, and a structure agnostic style vector produced by the style encoder. The mask estimator predicts a mask to blend the synthesized content into the original image at a proper location. In what follows, we describe each of these components in detail.

\noindent \textbf{Mask estimator.} The mask estimator $M$ is an encoder-decoder style network that maps an image $x$ and a partial sketch map $c$ to a mask $m$, \ie $m=M(x,c)$. The mask is produced by applying the sigmoid function on the output of a convolutional layer, which is a gray-scale map of range $[0,1]$. 
We use the term mask for consistency with previous inpainting-based approaches. 
Each element of the mask represents how likely the corresponding pixel of the input image should be modified to match the sketch. Thus, we can separate the modification region and non-modification static region by multiplying an input image $x$ with $m$ and $1-m$ element-wise, respectively. Since we are interested in the style information in the modification region, we define a style partial image as  $x^{sty}=m\odot x$, where $\odot$ represents element-wise multiplication. We define a static partial image  $x^{sta}=(m-1)\odot x$ as the non-modification region of the input image. 

% \begin{figure}[htbp]
% \begin{center}
% % \fbox{\rule{0pt}{1.5in} \rule{\linewidth}{0pt}}
% \includegraphics[width=.9\linewidth]{model_overview_short.pdf}
% \end{center}
% % \vspace{-10pt}
% \caption{Model structure of the proposed method. \zl{ZL: I feel it could be better by drawing it horizontally. Also, it is still better to have a bit more info on the caption.}}
% \label{fig_model}
% \end{figure}

\noindent \textbf{Structure agnostic style encoder}. 
The goal of the style encoder $S$ is to extract the style information from the modification region of the input image but get rid of the structure information. It is composed of stacked convolutional layers followed by a global max pooling, which produces a $d$-dimensional style vector $^{sty}$ given the style partial image $x^{sty}$ and the mask $m$,~\ie $v^{sty}=S(x^{sty},m)$. By repeating $v^{sty}$ by $h\times w$ times, we can obtain a style feature $\hat{v}^{sty}_{ (d\times h\times w)}$ of arbitrary spatial size $h\times w$ while keeping position invariant. However, it may still be sensitive to spatial transformation due to the local connectivity of convolutional layers. To further encourage the extracted style features to be structure agnostic, we apply random warping and deterioration to $x^{sty}$ in training, which will be introduced in Sec.~\ref{sec:learning}. 

\noindent \textbf{Generator.} 
The proposed framework is general and does not depend on the specific architecture of the generator. We can combine it with most existing inpainting-based models by using the estimated modification regions as missing regions and tiling the style vector into style feature maps of the suitable size as a generator's input. In our implementation, we incorporate the network architecture of DeepFill-v2~\cite{yu2019free} due to its simplicity and lightweight. The generator $G$ consists of a coarse stage $G_0$ and refinement stage network $G_1$. The coarse stage takes the static partial image, estimated mask, and sketch as input. 
We use the style feature maps $\hat{v}^{sty}$ as the side input at the bottleneck of the coarse stage network. The output of the coarse stage is denoted as $y_0$, which is passed through the refinement stage to produce a refined output $y_1$:
\begin{equation}
\begin{split}
y_0&=G_0(x^{sta}, m, \hat{v}^{sty})\\
y_1&=G(x^{sta},m,\hat{v}^{sty})=G_1(y_0).\\
\end{split}
\end{equation}
The final result $y$ is produced by blending the refinement stage output into the original image using the estimated mask
\begin{equation}
y = y_1\odot m + x\odot(1-m).
\end{equation}

To train the generator, we minimize the L1 error between the output $y_0, y_1, y$ and the original image $x$. We also use adversarial training with a discriminator to encourage visual realism of the blended image $y$. The detailed training strategy and loss functions will be described in the next section. 

\subsection{Learning by reconstruction}
\label{sec:learning}
% \begin{figure*}[t]
% \begin{center}
% % \fbox{\rule{0pt}{1.5in} \rule{\linewidth}{0pt}}
% \includegraphics[width=\linewidth]{fig_train.pdf}
% \end{center}
% % \vspace{-10pt}
% \caption{Illustration of a training iteration. }
% \label{fig_train}
% \end{figure*}
\begin{figure*}[t]
\begin{center}
% \fbox{\rule{0pt}{1.5in} \rule{\linewidth}{0pt}}
\includegraphics[width=.9\linewidth]{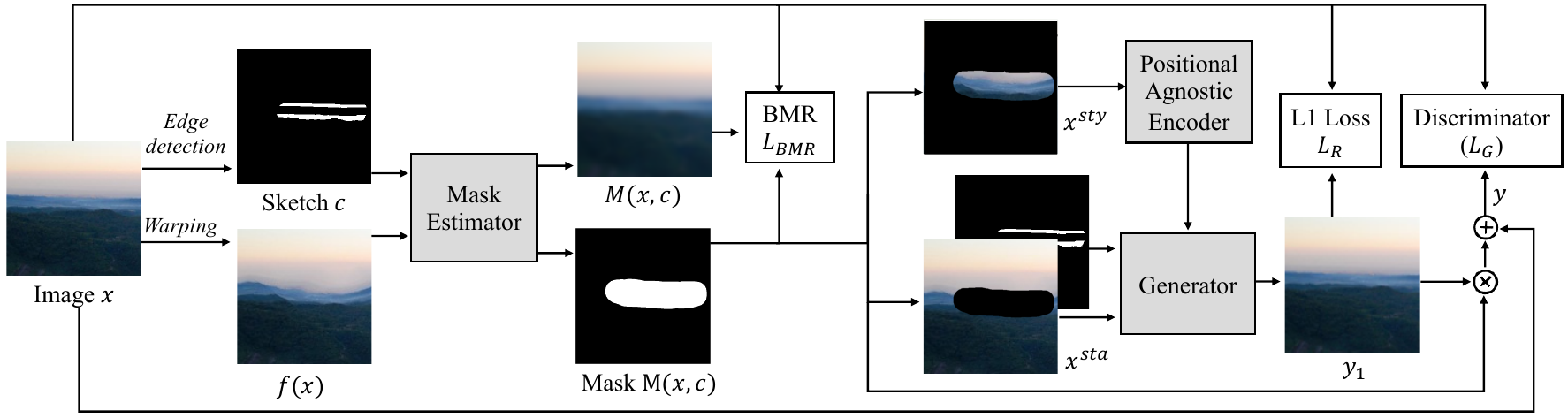}
\end{center}
% \vspace{-10pt}
\caption{Illustration of a training iteration. The mask estimator, style encoder, and generator are trained jointly by learning to reconstruct the original images.  To avoid the trivial solution of generating all-zero masks, we warp the input image in a local areal before passing them through the mask estimator. A bi-directional mask regularization loss (BMR) is applied to help the mask estimator produce reasonable masks rather than simply detecting warping artifacts. 
%\zl{ZL: a bit more info on caption.}
}
\label{fig_train}
\end{figure*}
%As it is too difficult to collect paired training data for sketch-based image manipulation, 
Due to the lack of training data, previous inpainting-based approaches resort to self-supervised learning, specifically, by learning to reconstruct the original image from a masked image and sketch. In this work, we also seek to train our model using self-supervision. However, as our model infers the mask, directly learning by reconstructing the original image will lead to a trivial solution. The mask estimator will constantly predict an all-zero mask; the generator becomes an identity mapping; the style encoder does not need to learn any meaningful representation.  
We solve this issue by deteriorating the input images with random local warping and regional dropout before passing them through the model. Fig.~\ref{fig_train} shows a training iteration of the proposed method. 
%To force the mask generator to predict the mask based on the sketch rather than simply detecting warping artifacts, we use a bi-directional mask regularization.
%for the mask estimator by making it estimate the original image from the warped ones and the warped images from the original ones. The region-wise dropout is applied on the style image $x^{sty}$ to further deviate its structure from the original image on top of warping, so the structure agnostic encoder can focus on extracting style information and ignoring the structure. 
%In what follow, we describe the training strategy and loss functions in details. 

\noindent \textbf{Local warping.} 
Before passing an input image through the model in training, we warp $x$ in a random area using triangular local warping. %As illustrated in Fig.~\ref{fig_warp}, 
We construct a triangular mesh by placing vertices on the image boundary, boundary of a randomly sampled region, and inside the region. Then a warping flow can be created by moving interior vertices while keeping other vertices unmoved. For general images, we place and move the interior vertices randomly. For face images, to make the warped faces remain visually plausible, we use facial landmarks and blendshapes to guide the selection and movements of the interior vertices. The detailed warping algorithm can be found in the supplementary material.  Fig.~\ref{fig_warp_image} shows samples augmented by warping. We can see that the warped images have the same color and texture as the original ones but with different structures in warped regions. Therefore, we can construct the original image with structure information from the sketch and style information extracted from a warped image.  Also, as a warped image is identical to the original image in non-warped regions, the mask estimator will be encouraged to predict zero where no sketch appears. Thus the regions irrelevant to sketches can be preserved. 
%%
% \begin{figure}[h]
% \begin{center}
% % \fbox{\rule{0pt}{1.5in} \rule{\linewidth}{0pt}}
% \includegraphics[width=\linewidth]{fig_warp_short.pdf}\\
% \scriptsize{\hfill{Original Image} \hfill\hfill  {Construct Mesh} \hfill\hfill  {Move Vertices} \hfill\hfill {Warped Image} \hfill\hfill}
% \end{center}
% % \vspace{-10pt}
% \caption{Illustration of local triangular warping. }
% \label{fig_warp}
% \end{figure}
%%
\begin{figure}[t]
\begin{center}
% \fbox{\rule{0pt}{1.5in} \rule{\linewidth}{0pt}}
%\includegraphics[width=\linewidth]{warped_image_edge.pdf}\\
%\scriptsize{\hfill{\textcolor{white}{war}Image\textcolor{white}{ped }} \hfill\hfill {\textcolor{white}{war}Sketch\textcolor{white}{ped }} \hfill\hfill  {Warped Image} \hfill\hfill {Warped Sketch} \hfill\hfill}
\includegraphics[width=\linewidth]{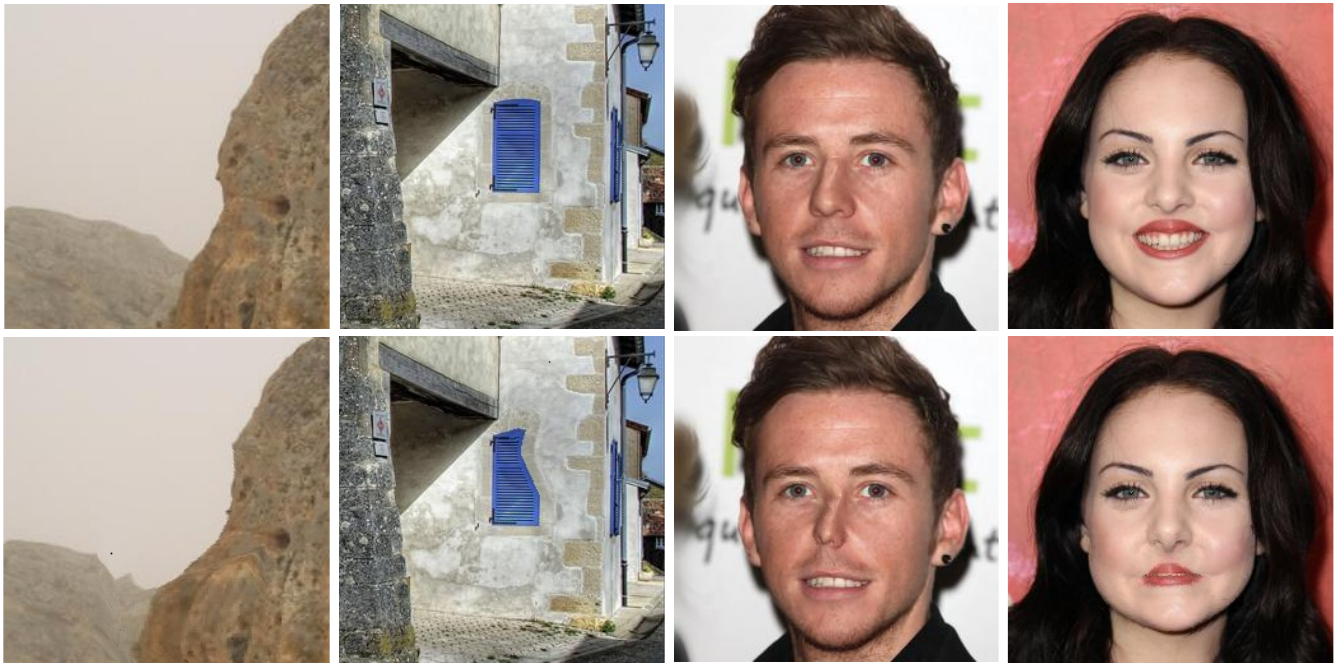}\\
\end{center}
% \vspace{-10pt}
\caption{Example images augmented by local warping. Top: original images, bottom: warped images. }
\label{fig_warp_image}
\end{figure}

\begin{figure}[t]
\begin{center}
% \fbox{\rule{0pt}{1.5in} \rule{\linewidth}{0pt}}
\includegraphics[width=\linewidth]{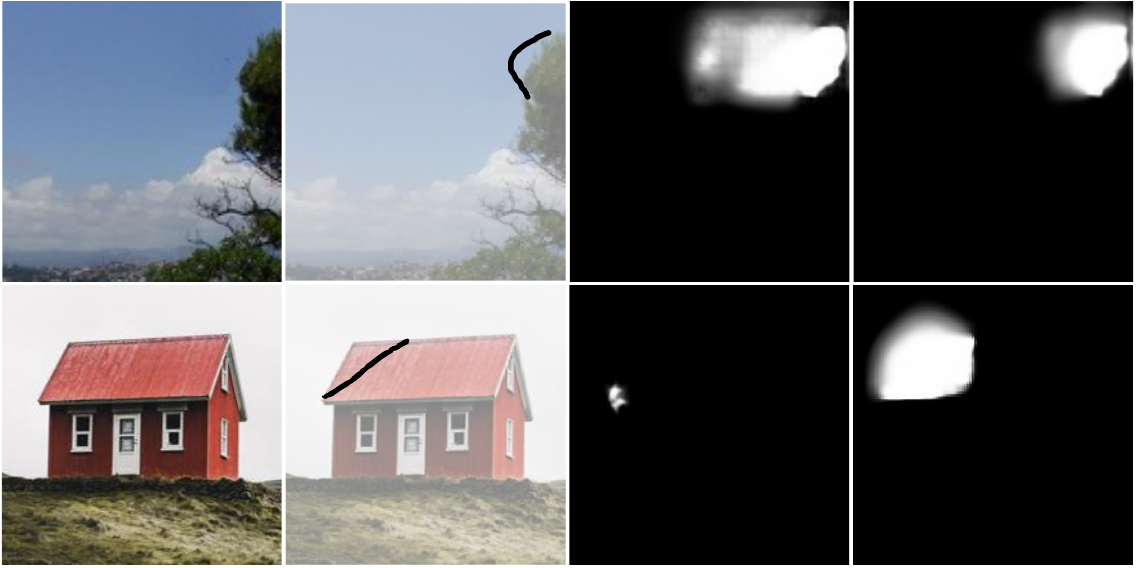}\\
\scriptsize{\hfill{Image} \hfill\hfill  {Sketch} \hfill\hfill  {w/o BMR} \hfill\hfill {w/ BMR} \hfill\hfill}
\end{center}
% \vspace{-10pt}
\caption{The predicted masks with and without the bi-directional mask regularization (BMR). Top: warped images, bottom: real images. BMR helps the mask estimator generalize better to real image inputs.}
\label{fig_cmp_mask2way}
\end{figure}

\noindent \textbf{Bi-directional mask regularization.} 
Since we warp the input images in training, the mask predictor may end up producing masks by simply detecting warping artifacts and fail in the inference stage, as shown in thee third column of Fig.~\ref{fig_cmp_mask2way}. To avoid this behavior, we train the mask estimator with a bi-directional image reconstruction loss as regularization. 
Let $f(x)$ denote the image produced by distorting $x$ with the warping flow $f$. By applying the same warping flow to the sketch $c$, we obtain a sketch $f(c)$ corresponding to the warped image. 
During training, we add an extra decoder to the mask estimator to predict an image, with the goal of estimating the original image $x$ from the warped image $f(x)$ and the original sketch $c$, and the opposite, to estimate $f(x)$ from $x$ and $f(c)$. Thus the mask estimator $M$ has an extra output in addition to the mask output $m=M(x,c)$. For input image $x$ and sketch $c$, we let $\bar{M}(x,c)$ represent the estimated image and $\hat{M}(x,c)$ represent the image blended using the mask,~\ie $\hat{M}(x,c)=\bar{M}(x,c)\odot M(x,c)+x\odot (1-M(x,c))$. 
The bi-directional mask regularization (BMR) loss is defined as follows:
\begin{equation}
\label{loss_mask}
\begin{split}
\nonumber L_{BMR} &= \norm{\bar{M}(f(x),c)-x}_1+\norm{\bar{M}(x,f(c))-f(x)}_1\\
    &+\norm{\hat{M}(f(x),c)-x}_1+\norm{\hat{M}(x,f(c))-f(x)}_1.
\end{split}
\end{equation}
Since $L_{BMR}$ is computed for both the warped and the original images, the mask estimator can be adapted to natural image input and thus can predict reasonable masks in the inference stage, as shown in  the forth column of Fig.~\ref{fig_cmp_mask2way}. 

\noindent \textbf{Regional dropout.} 
Warping helps the style encoder learn structure agnostic representation by creating structural variation while preserving the style in a local region. However, the warped image and the original image usually have identical visual elements, while real cases might involve the generation of new elements. Therefore, during training, we randomly dropout sampled regions in the style image $x^{sty}$ before passing it through the style encoder to reduce the correlation between $v^{sty}$ and $x$. 

\noindent \textbf{Loss functions.} For a pair of image $x$ and sketch map $c$ in the training set, we compute the following loss functions to jointly train the mask estimator, style encoder, and the generator:
\begin{equation}
\label{loss_l1}
L_R = \norm{y_0-x}_1 + \norm{y_1-x}_1 + \norm{y-x}_1.
\end{equation}
Eqn.~\ref{loss_l1} measures the error of the coarse stage output $y_0$ and refinement stage output $y_1$, as well as the error of the manipulated image $y$, which is the combination of $x$ and $y_1$ blended using the estimated mask.  In addition,
\begin{equation}
\label{loss_g}
L_G =\mbox{ReLU}[\mathbbm{1}-D(y, c)],
\end{equation}
where $D$ represents the discriminator. For the discriminator, we use the hinge loss and spectral normalization~\cite{miyato2018spectral} in~\cite{yu2018generative,yu2019free}. 
Eqn.~\ref{loss_g} defines the adversarial loss inspired by GANs. To minimize $L_G$, the generator and mask estimator have to cooperate reasonably to fool the discriminator. It requires the generator to synthesize visually realistic content and the mask estimator to provide the mask that blends the synthesized content into a proper region in the original image seamlessly. The overall loss function is the sum of $L_G$, $L_R$ and the mask regularization term $L_{BMR}$:
\begin{equation}
\label{loss_total}
L_{total} = L_R+L_G+L_{BMR}.
\end{equation}

\section{Experiments}
%\subsection{Implementation Details}
%We implemented our method using Python and Pytorch. The network is optimized with the Adam~\cite{kingma2014adam} optimizer with the learning rate set equal to $1e-4$. We use a cloud server with 4 NVIDIA A100 GPUs for training. It takes about $x$ seconds to process an image at $256\times 256$ resolution and $y$ seconds at $512\times 512$ resolution on GPU. The detailed network architectures, code and model can be found in the supplementary material. 
As introduced in Sec.~\ref{sec:model}, our framework can be combined with most inpainting-based models to enable mask-free manipulation. In our implementation, we plug the DeepFill-v2 generator into our framework due to its simplicity and efficiency. 
To verify the effectiveness of the proposed method, we compare the results under our mask-free framework to the original DeepFill-v2 as a baseline. We also compare with other state-of-the-art inpainting-based face manipulation methods including SC-FEGAN~\cite{jo2019sc} and DeepPS~\cite{yang2020deep} to demonstrate the superiority of our approach. 

\subsection{Datasets \& Evaluation Metric}
We train our general image manipulation model on the training split of Places~\cite{simonyan2014very}. For face manipulation, we train the model on  CelebHQ~\cite{karras2017progressive} for consistency with the inpainting-based approaches. We randomly sample 2,000 images as the validation set and use the remaining data for training. 

For evaluation, we use edge maps extracted by~\cite{he2019bi} as surrogate sketches for Places and obtain the sketches in face regions by connecting detected facial landmarks for the CelebAHQ dataset. We randomly sample modification regions for each test image and obtain partial sketches by discarding all edges outside this region. For evaluation of our method, we apply local warping in modification regions. For inpainting-based methods, we set pixel values in modification regions equal to zero according to their training condition. 
% \textcolor{red}{For face images where the modification regions are sampled as facial components, we obtain the region masks by the minimum bounding box enclosing the original and warped partial sketches. }
For synthetic samples, the original images can be seen as the ground-truth. Hence, we evaluate the performance using PSNR, L1 error, SSIM, and FID~\cite{heusel2017gans} between the results and the original images. PSNR, L1 error and SSIM are based on pixel-wise errors, often used in image inpainting and restoration tasks~\cite{yu2018generative,dong2015image,zhang2019residual,han2018image,xie2021exploiting,wang2019spatial,ren2019progressive,li2018recurrent}. 
FID measures the distance between the distribution of the deep features of generated images and real images. It is the current standard metric for assessing the quality of generative models~\cite{lucic2018gans}. 

Apart from using synthetic samples for evaluation, to reflect the performance in real cases, we construct two test sets SketchImg and SketchFace for face and general image manipulation, respectively. These two datasets contain 100 natural images and 400 face images as well as the corresponding sketches and masks drawn by human users. 
The samples are created under the conditional inpainting framework, where the users draw a sketch indicating the target contour and the mask that marks the region to modify. To evaluate our method, we simply ignore the masks and only use the images and sketches as input. For the real cases, the results are supposed to have a similar style to the input images. Therefore, we evaluate the performance using style loss~\cite{gatys2015neural},~\ie the mean squared error between the Gram matrices of the deep features of the input image and results. We use the feature maps from the \texttt{relu\_1} and \texttt{relu\_2} of a VGG~\cite{simonyan2014very} network pretrained on ImageNet~\cite{deng2009imagenet}. We denote the obtained style loss as SL\texttt{\_1} and SL\texttt{\_1}, respectively. We also report FID between the results and input images as a reflection of the image quality. 

\subsection{Qualitative Evaluation} 
\begin{figure*}[h]
\begin{center}
%\fbox{\rule{0pt}{2in} \rule{\linewidth}{0pt}}
\includegraphics[width=\linewidth]{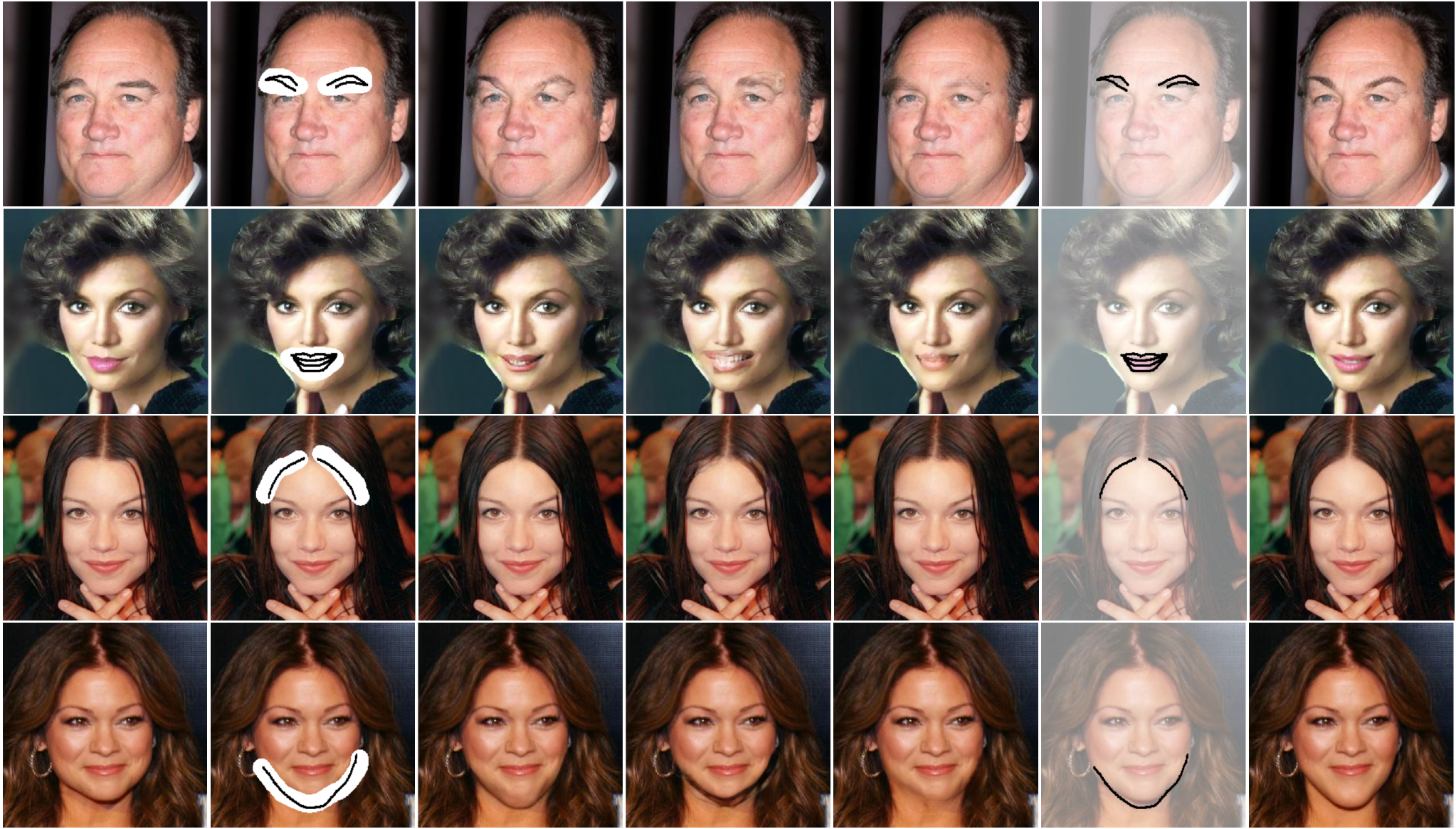}\\
\scriptsize{\hfill {\textcolor{white}{aaa}Image\textcolor{white}{aa}} \hfill\hfill  {\textcolor{white}{aa}Input*\textcolor{white}{aa}} \hfill\hfill  {DeepFill-v2} \hfill\hfill {\textcolor{white}{a}SC-FEGAN\textcolor{white}{a}} \hfill\hfill {\textcolor{white}{aa}DeepPS\textcolor{white}{aa}} \hfill\hfill {\textcolor{white}{aaa}Input\textcolor{white}{aa}} \hfill\hfill {\textcolor{white}{aaa}Ours\textcolor{white}{aaa}} \hfill\hfill}
\end{center}
\vskip-8pt\caption{Visual comparison of face manipulation results on CelebAHQ dataset. Input*: input sketch and masks for inpainting-based methods. Input: input sketch for our method. }
\label{fig_cmp_celeb}
% \vspace{-20pt}
\end{figure*}

\begin{figure*}[h]
\begin{center}
%\fbox{\rule{0pt}{2in} \rule{\linewidth}{0pt}}
\begin{minipage}{\textwidth}
\includegraphics[width=\linewidth]{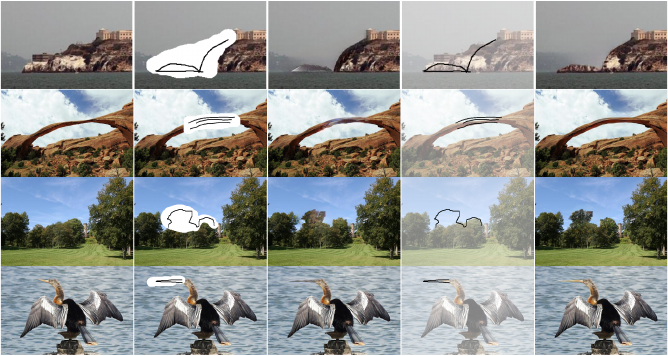}\\
\scriptsize{\textcolor{white}{a}\hfill{\textcolor{white}{aa}Image\textcolor{white}{aa}} \hfill\hfill  {\textcolor{white}{aaa}Input*\textcolor{white}{aa}} \hfill\hfill  {DeepFill-v2} \hfill\hfill {\textcolor{white}{aaa}Input\textcolor{white}{aa}} \hfill\hfill {\textcolor{white}{aaa}Ours\textcolor{white}{aaa}}\hfill}
\end{minipage}
\end{center}
\vskip-8pt\caption{Visual comparison of general image manipulation results on the Places dataset. Input*: input sketch and masks for inpainting-based methods. Input: input sketch for our method. }
\label{fig_cmp_places}
% \vspace{-20pt}
\end{figure*}

Fig.~\ref{fig_cmp_celeb} presents the face manipulation results of our method and the state-of-the-art methods. 
As our model leverages entire images as input, it can better reserve the original appearance in the modification region of the input image,~\eg the eyebrows in the firs row and the lip color in the second row. 
% \textcolor{red}{Also, it can be seen that all previous approaches take a sketch map and a mask as input, while our method only needs the sketch. Therefore, our method provides a simpler user interface and is less sensitive to users' behavior. As shown in the third and the fourth row, previous approaches generate different results with different sizes of masks for the same sketch input, while our method can produce consistent results. }
Our method can also produces visually realistic results for more challenging general image manipulation. 
Fig.~\ref{fig_cmp_places} shows the general image manipulation results of our method and DeepFill-v2. 
It can be seen that our method well preserves the colors and textures of the original images. In contrast, DeepFill-v2 often adds extra changes in the modification regions,~\eg it changes the colors of the island, arch, and the beak in the first, second, and forth row, respectively. 
In addition, as our framework prevents the information loss caused by masking out the modification area, our results are less blurry and have more details. 


\subsection{Quantitative Evaluation}
Table~\ref{table_cmp_syn} reports the quantitative evaluation results on the synthetic samples from the CelebAHQ and Places validation sets. It can be seen that our results have larger PSNR, SSIM and smaller L1 error than previous approaches, which imply a smaller difference between the generated images and the corresponding ground-truth. Our results also have smaller FID, indicating a closer distance to the distribution of original images. 
Table~\ref{table_cmp_real} reports the quantitative evaluation results on real samples. It can be seen that the style loss of our results are significantly smaller than previous approaches, as our mask-free framework effectively keeps the style of the input images. Our method also outperforms previous approaches in terms of FID on real samples. 

\begin{table}[h]
\caption{\small Quantitative comparison on synthetic samples from CelebAHQ (top) and Places (bottom) validation sets. }
\begin{center}
\label{table_cmp_syn}
\small
\begin{tabular}{ccccc}
\hline
\multicolumn{5}{c}{CelebAHQ}\\
Method & L1 error$\downarrow$ & PSNR$\uparrow$& SSIM$\uparrow$ & FID$\downarrow$\\
\hline
SC-FEGAN & 0.0110 & 30.39 & 0.9408 &2.693\\
% SC-FEGAN (full edge) & 0.0111 & 30.09 & 0.9402 & x\\
DeepPS & 0.0125 & 30.54 & 0.9340 & 2.460\\
% DeepPS (full edge) & 0.0125 & 30.52 & 0.9338 \\
DeepFill-v2 &0.0115 & 30.59 & 0.9370 &2.345\\
% DeepFill-v2 (my)(full edge) &0.0090 & 32.47 & 0.9507 &x \\
% Ours(v1) & 0.0069 & 35.40 & 0.9688 &0.721\\
Ours &\textbf{0.0071} & \textbf{34.21} & \textbf{0.9623} &\textbf{0.806}\\
% \hline
\end{tabular}
\small
\begin{tabular}{ccccc}
\hline
\multicolumn{5}{c}{Places}\\
Method & L1 error$\downarrow$ & PSNR$\uparrow$ & SSIM$\uparrow$ & FID$\downarrow$\\
\hline
DeepFill-v2 &0.0293 & 24.06 & 0.8442 &2.238\\
% Ours(v1) & 0.0202 & 26.81 & 0.8681 &1.305\\
Ours &\textbf{0.0223} & \textbf{26.35} & \textbf{0.8527} & \textbf{1.500}\\
\hline
\end{tabular}
\vspace{-10pt}
\end{center}
\end{table}
\begin{table}[h]
\setlength{\tabcolsep}{1.5pt}
\caption{\small Quantitative comparison on SketchFace and SketchIMG containing sketches and masks drawn by human users.}
\begin{center}
\label{table_cmp_real}
\small
\resizebox{\columnwidth}{!}{
\begin{tabular}{ccccc|cc}
\hline
\multicolumn{5}{c|}{SketchFace}&\multicolumn{2}{c}{SketchImg}\\
% \hline
Method & SC-FEGAN & DeepPS & DeepFill-v2 & Ours & DeepFill-v2 & Ours\\
\hline
FID$\downarrow$ & 4.092 & 5.285 &4.382 &\textbf{3.31} &27.11 &\textbf{24.27}\\
SL\texttt{\_1}$\downarrow$ & 2484 & 3797 & 2859 & \textbf{2211} & 5314 & \textbf{4374}\\
SL\texttt{\_2}$\downarrow$ & 13145 & 15585 & 13970 & \textbf{10819} & 24169 & \textbf{21575}\\
\hline
\end{tabular}
}
\vspace{-10pt}
\end{center}
\end{table}

\subsection{Ablation Study}
\begin{figure}[t]
\begin{center}
%\fbox{\rule{0pt}{2in} \rule{\linewidth}{0pt}}
\includegraphics[width=\linewidth]{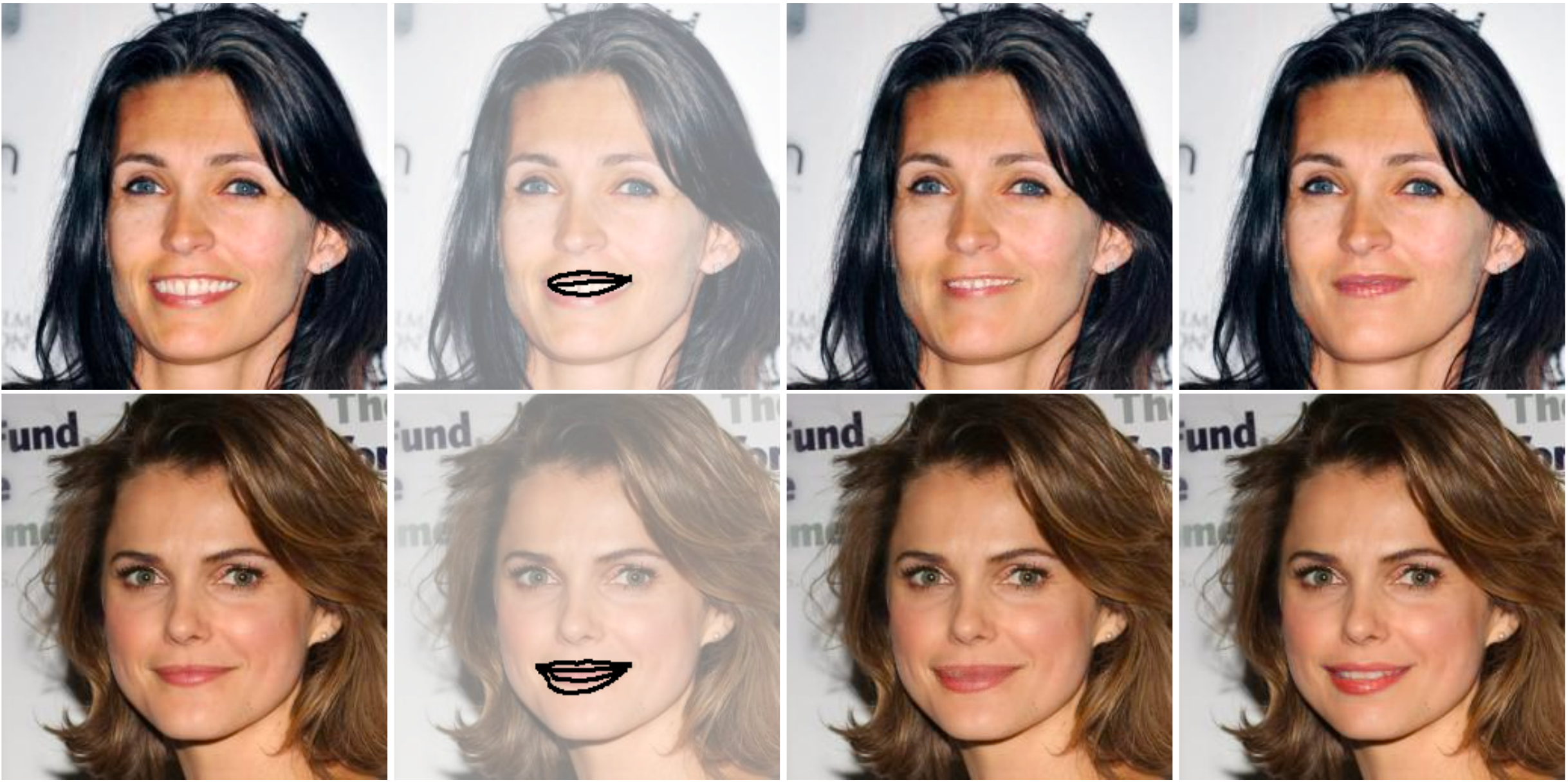}\\
\small{\hfill{Sketch} \hfill\hfill  {Image} \hfill\hfill  {w/o ME} \hfill\hfill {with ME} \hfill\hfill}
\end{center}
\vskip-8pt\caption{Top: results of the model without the mask estimator (ME), bottom: results with ME.}% The mask estimator improves generative capability of the model. }
\label{fig_abl_mask}
% \vspace{-20pt}
\end{figure}

\noindent \textbf{Mask estimator.}  
The mask estimator is an important component in our framework to enable local manipulation and encourage the model to learn better generative capability. 
Without the mask predictor, we may train a common conditional GAN on training samples synthesized by warping. However, a model trained to mimic warping often fails in challenging cases involving generating new elements or deleting existing elements. For example, as shown in Fig.~\ref{fig_abl_mask}, the generator jointly trained with the mask estimator correctly deletes and adds teeth in manipulation of mouths. In contrast, a common conditional GAN only squeezes or stretches the mouth. 
The mask estimator also helps to better preserve the non-modification regions. Without the mask estimator, slight changes in the static region is almost inevitable. This is reflected by the larger L1 error and smaller PSNR in the third row of Table~\ref{table_abl_places}. 
Another baseline without the mask estimator is to use masks generated based on rules. The third row reports the results with the rule-based masks given by the minimum bounding boxes enclosing sketches. By comparing the first row and the third row, we can see the significant improvements brought by the mask estimator. 
% The fourth row shows the style loss obtained with trivial masks generated by dilating the sketches. We can see that trivial masks only slightly help. In comparison, the mask estimator can significantly improve the quality of the results by predicting adaptive masks that add only necessary modifications to the input. 
\begin{table}[h]
\setlength{\tabcolsep}{3pt}
\caption{\small Effect of each components. Rule-based Mask: masks given by the minimum bounding boxes enclosing sketches, ME: mask estimator, SE: style encoder. The first row corresponds to the full model with both ME and SE. }
\begin{center}
\label{table_abl_places}
\small
\begin{tabular}{cccccc}
\hline
% w/o Mask & Trivial Mask & ME & SE &SL\texttt{\_1}\\% &SL\texttt{\_2} \\
% \hline
%   & & \checkmark & \checkmark &2211 \\%& 0.0071& 34.21&0.9623\\\\
%   & & \checkmark &            &2463 \\%&0.0073 & 34.15 & 0.9627\\
% \checkmark &   &  & \checkmark &6105 \\%&0.0127 & 32.50 & 0.9499\\
%   & \checkmark & & \checkmark &x \\%\\&x\\
% \hline
w/o Mask & Rule-based Mask & ME & SE &L1 error$\downarrow$ &PSNR$\uparrow$\\% &SL\texttt{\_2} \\
\hline
      & & \checkmark & \checkmark & 0.0071& 34.21\\%&0.9623\\\\\
   \checkmark &   &  & \checkmark &0.0127 & 32.50 \\%& 0.9499\\
   & \checkmark & & \checkmark&0.0102 & 31.87 \\%\\&x\\
      & & \checkmark &            &0.0074 & 33.98 \\%& 0.9627\\
\hline
\end{tabular}
% \begin{tabular}{ccccc}
% \hline
% w/o Mask & Trivial Mask & ME & SE &FID \\
% \hline
%   & & \checkmark & \checkmark &x\\
%   & & \checkmark &            &x\\
% \checkmark &   &  & \checkmark &x\\
%   & \checkmark & & \checkmark &x\\
% \hline
% \end{tabular}
\vspace{-10pt}
\end{center}
\end{table}

\begin{figure}[t]
\begin{center}
%\fbox{\rule{0pt}{2in} \rule{\linewidth}{0pt}}
\includegraphics[width=\linewidth]{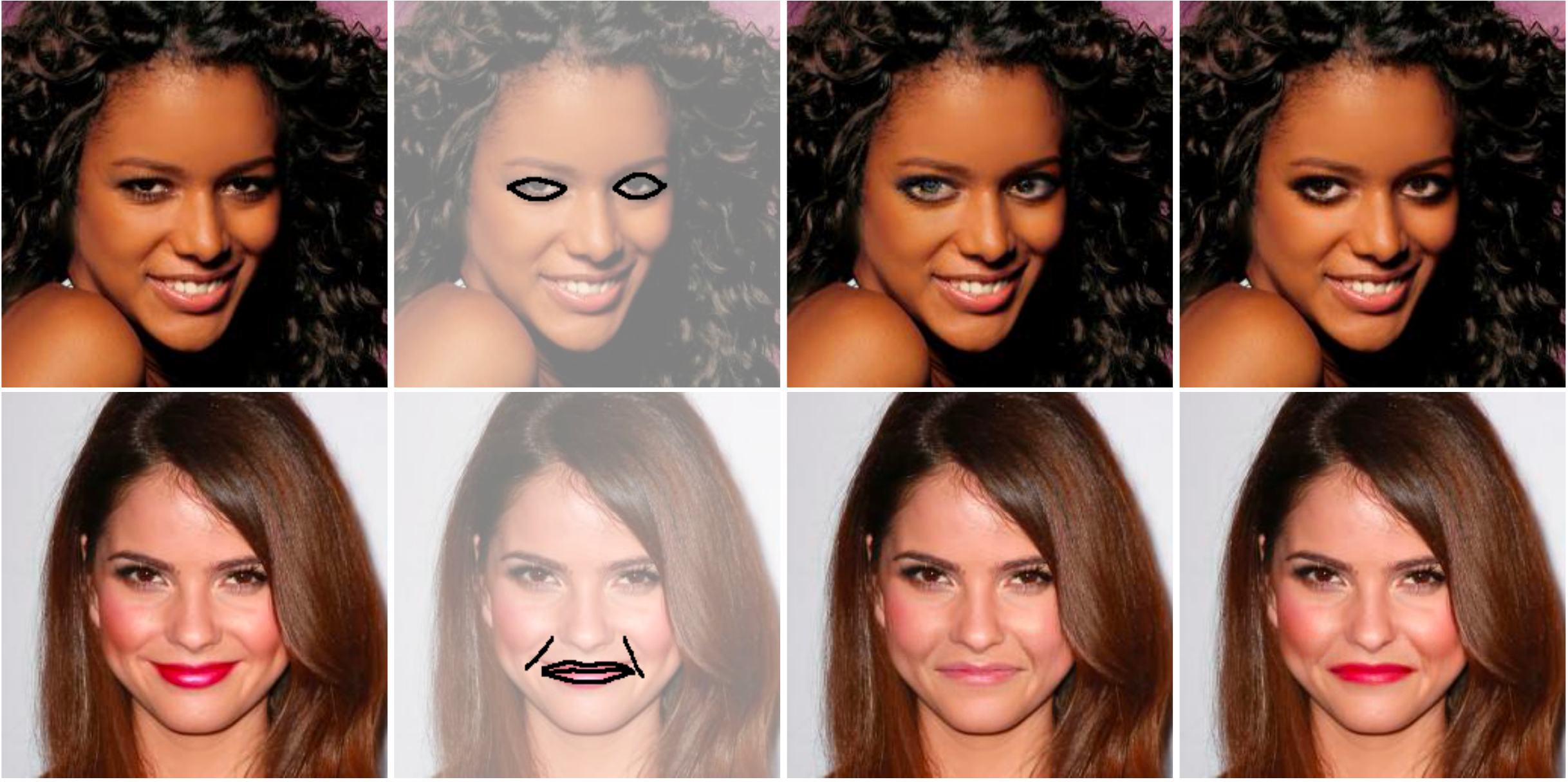}\\
\small{\hfill{Sketch} \hfill\hfill  {Image} \hfill\hfill  {w/o SE} \hfill\hfill {with SE} \hfill\hfill}
\end{center}
\vskip-8pt\caption{Top: results of the model without the style encoder (SE), bottom: results with SE. }%The style encoder (SE) preserves the appearance of the modification region. }
\label{fig_abl_style}
% \vspace{-20pt}
\end{figure}

\noindent \textbf{Style encoder.}  The style encoder is essential to keep the appearance of the modification region. The model without the style encoder has no access to information in the modification region and generates new content based on prior, resulting in unexpected changes. Fig.~\ref{fig_abl_style} shows example results for face manipulation. It can be seen that the model with the style encoder reserves the colors of the lips and eyes, while the one without the style encoder changes the colors. 
The quantitative evaluation results also reflect the effect of the style encoder. The first and the forth row of Table~\ref{table_abl_places} report the quantitative evaluation results with and without the style encoder, respectively. It can be seen that the model with the style encoder results in smaller L1 error and larger PSNR.

\section{Conclusion, Limitations and Future Work}
\label{sec:conc}
This paper presents a new paradigm of sketch-based image manipulation,~\ie mask-free local image manipulation, and propose a complete system implementation for this task including the network architecture, data acquisition and training strategy. The proposed method offers simpler and more robust user workflows for sketch-based image manipulation and provides consistently better results than the related approaches. A limitation is that the current framework only supports structure editing based on monochrome sketches. Color editing based on colored sketches under the proposed new mask-free framework is an interesting topic for future work. 
Potential negative social impact of this work might include: fake news can be created by manipulating photos; The spreading of manipulated images may distort of people's perception of body images and beauty. The negative impact can be mitigated by embedding digital watermarks in manipulated images to make them easy to identify.

{\small
\bibliographystyle{ieee_fullname}
\bibliography{egbib}
}

\end{document}